\pdfoutput=1

\documentclass[11pt]{article}
\usepackage{listings}
\usepackage{graphicx}
\usepackage{ACL2023}
\usepackage{xcolor}
\usepackage{times}
\usepackage{latexsym}

\usepackage[T1]{fontenc}

\usepackage[utf8]{inputenc}

\usepackage{microtype}

\usepackage{inconsolata}
\usepackage{xcolor}
\usepackage{fdsymbol}

\lstdefinestyle{python}{
  language=Python,
  basicstyle=\ttfamily\small,
  keywordstyle=\color{blue},
  commentstyle=\color{gray},
  stringstyle=\color{red},
  showspaces=false,
  showstringspaces=false,
  tabsize=4,
  breaklines=true,
  aboveskip=-5pt,
  belowskip=-7pt
}

\title{Explaining Competitive-Level Programming Solutions using LLMs}

\author{Jierui Li $^{\varheartsuit}$, Szymon Tworkowski$^{\clubsuit,\spadesuit}$, Yingying Wu$^{\vardiamondsuit}$ and Raymond Mooney$^{\varheartsuit}$ \\
$^{\varheartsuit}$The University of Texas at Austin,  ~$^{\clubsuit}$University of Warsaw, $^{\spadesuit}$IDEAS NCBR \\
$^{\vardiamondsuit}$ University of Houston\\
\texttt{\{jierui, mooney\}@cs.utexas.edu}, \texttt{szy.tworkowski@gmail.com}, \texttt{ywu68@uh.edu} }

\begin{document}
\maketitle
\begin{abstract}
 In this paper, we approach competitive-level programming problem-solving as a composite task of reasoning and code generation. We propose a novel method to automatically annotate natural language explanations to \textit{<problem, solution>} pairs. We show that despite poor performance in solving competitive-level programming problems, state-of-the-art LLMs exhibit a strong capacity in describing and explaining solutions. 
 Our explanation generation methodology can generate a structured solution explanation for the problem containing descriptions and analysis. To evaluate the quality of the annotated explanations, we examine their effectiveness in two aspects: 1) satisfying the human programming expert who authored the oracle solution, 
 and 2) aiding LLMs in solving problems more effectively. 
The experimental results on the CodeContests dataset demonstrate that while LLM GPT3.5's and GPT-4's abilities in describing the solution are comparable, GPT-4 shows a better understanding of the key idea behind the solution. 
\end{abstract}

\section{Introduction}

Recent Large Language Models (LLMs) have shown impressive capabilities for various reasoning tasks, including multi-hop question answering \cite{wang-etal-2022-iteratively, lyu2023faithful}, commonsense reasoning \cite{zelikman2022star}, symbolic reasoning \cite{hua-zhang-2022-system}, and math word problem-solving \cite{ zhou2023leasttomost, chen2022program}. The chain-of-thought prompting method \cite{wei2022chain} explicitly instructs LLMs to generate intermediate steps until the final answer is reached, enabling the model to decompose problems and solve them step-by-step. Nevertheless, challenges persist when tackling complex reasoning tasks, such as competitive-level programming problems. For instance, even powerful models like GPT-4 outperform fewer than 5\% of human competitors in virtual contests from Codeforces \cite{openai2023gpt4}. Competitive-level programming problems epitomize problem-solving strategies for algorithmic, mathematical, geometric, and graph-theoretic problems. Solving them necessitates understanding problems, familiarity with algorithms, reasoning skills, creative algorithm development, and efficient, robust implementation.
\begin{table}[ht]
  \centering
    \begin{tabular}{|p{7.3cm}|}
    \hline
   \small \textbf{Problem Statement: Sign Swap} \\
    \small Given an array of $n$ integers $a_1, a_2, \ldots, a_n$, where $a_i \neq 0$, check if you can make this array sorted by using the following operation any number of times (possibly zero). An array is sorted if its elements are arranged in a non-decreasing order. Select two indices $i$ and $j$ $(1 \leq i, j \leq n)$ such that $a_i$ and $a_j$ have different signs. In other words, one must be positive and one must be negative. Swap the signs of $a_i$ and $a_j$. For example if you select $a_i=3$ and $a_j=-2$, then they will change to $a_i=-3$ and $a_j=2$.
 \\ \hline
    \small \textbf{Solution:} \\ 
    \vspace*{-2mm}
    \begin{lstlisting}[style=python,basicstyle=\small]
a= [*map(int,input().split())]
h=sum(1 for v in a if v<0)
b=[abs(a[i]) * (-1 if i<h else 1) for i in range(len(a))]
print('yes' if sorted(b)==b else 'no')
    \end{lstlisting}
    \vspace*{-3mm}\\ \hline
    \small \textbf{Solution Description:} Move negative signs to the front of the array and check if it's already non-decreasing. \\
    \small \textbf{Solution Explanation:} Swapping different signs for any number of times means you can move negative signs arbitrarily. A non-decreasing array must have negative elements ahead of positive elements and moving negative signs ahead is the optimal operation can be made.\\ \hline
    \small \textbf{GPT-3.5 Generated Program: }\\
    \vspace*{-2mm}
    \begin{lstlisting}[style=python, basicstyle=\tiny]
a = list(map(int, input().split()))
neg_count = sum(1 for x in a if x < 0)
pos_count = len(a) - neg_count
while True:
    if neg_count > pos_count:
        for i in range(len(a)):
            if a[i] < 0: a[i] = -a[i]; neg_count -= 1; pos_count += 1; break
    elif pos_count > neg_count:
        for i in range(len(a)):
            if a[i] > 0: a[i] = -a[i]; pos_count -= 1; neg_count += 1; break
    else: break
if all(a[i] <= a[i+1] for i in range(len(a)-1)): print("YES")
else: print("NO")
    \end{lstlisting}
    \vspace*{-3mm}\\ \hline
  \end{tabular}
  \caption{An example of a problem and solution from Codeforces with human oracle Solution, Description, and Explanation. The formatting in the problem is simplified. Program generated by GPT-3.5 is incorrect.}
  \label{tab:examples}
\end{table}

Previous works on automatically solving programming problems focus on tasks mapping fairly detailed natural language instructions to programs. \citet{Li_2022,ni2023lever,chen2022codet} verified and selected candidate programs by running them on human-written or automatically generated test cases. \citet{shojaee2023executionbased, chen2023teaching, schäfer2023adaptive} incorporated execution feedback as an intermediate step during code generation to enhance the programming ability of LLMs. While these methods yield promising results for fairly straightforward implementation tasks, they fall short on algorithmic reasoning tasks. 

Table \ref{tab:examples} shows a sample problem from Codeforces.\footnote{\url{https://codeforces.com/problemset/problem/1670/A}} Compared to most instruction-to-program tasks, competitive-level programming, a problem-to-program task, is more challenging. Before implementing the program, one needs to first abstract the problem and create a mathematical representation, come up with potential solutions, consider time and space complexity constraints, and corner cases, and finally identify a proper problem-solving strategy.

In order to disentangle reasoning about the problem from code implementation, we advocate decomposing the process of solving a programming problem. Instead of directly generating a program given the problem statement, we propose adding explicit, intermediate reasoning steps in natural language. These steps are more aligned with how humans typically solve such problems and utilize the idea of chain-of-thought. 

\begin{figure*}
	\centering
		\includegraphics[width=0.82\textwidth]{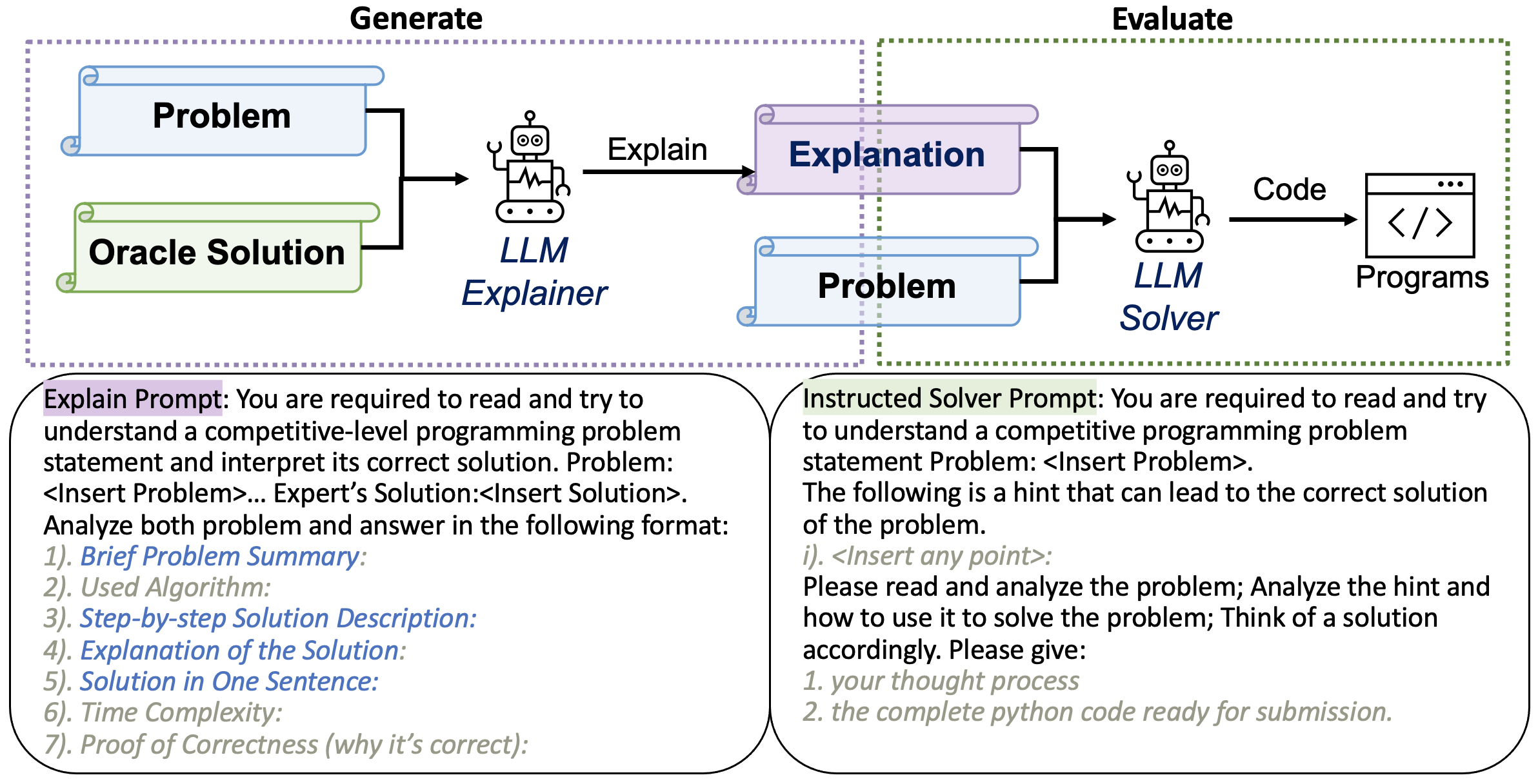}
	\caption{The explanation generation and evaluation framework and corresponding prompts (Top). An example of the full explain prompt (Bottom Left) and model's output is in Appendix Table \ref{tab:case_study}. The blue points are descriptions while the grey points are analysis. We give the explanation based on the oracle solution to the instructed solver as a hint (Bottom Right) to evaluate the quality of the generated explanation. }
	\label{fig:main method}
\end{figure*}

However, while \textit{<problem, solution>}\footnote{A solution here refers to a correct program.} pairs are publicly available on the practice websites \cite{Li_2022}, natural language descriptions or explanations of ``how to solve the problem'' are hard to collect at scale, as it requires additional annotations from programming experts. 
Hence, we propose to automatically generate explanations using an LLM. We found that: 1) Given both the problem and a human-written program solution,

LLMs like GPT-3.5 and GPT-4 can describe the solution in natural language reasonably well; and 2) Given the automatically generated explanation as a hint, LLMs perform better at solving the problem. 

Our explanation generation methodology shown in Figure \ref{fig:main method} employs a hierarchical prompt, requesting a step-by-step explanation from detailed code analysis to a broader understanding of the solution. The 7 points can be categorized into \textit{Description-Level} (i.e., Point 1,3,4,5) and \textit{Analysis-Level} (i.e., Point 2,6,7) of the problem and solution. To evaluate the quality of the generated explanations, we examine their effectiveness in two respects: 1) being positively rated by the human programming expert who authored the ``oracle'' solution, and 2) aiding LLMs in solving the problems more effectively. In the explanation-aided evaluation, Explainer generates the explanation given the oracle solution while Solver generates programs from the explanation. We use GPT-turbo-3.5 and GPT-4 as the Explainer and GPT-turbo-3.5 as the Solver in our experiments. 

In the human evaluation, we ask human experts who are the authors of solutions to score the explanations from -2 to 2. They give the explanations positive scores averaging $0.81$ and $1.30$ on GPT-3.5 Explainer and GPT-4 Explainer respectively.  With respect to explanation-aided program synthesis, we find that different points of the generated explanations can guide the model to solve the problem better, with the solve rate at pass@10 (one of the top 10 generated programs is deemed correct) increasing from 6.1\% to 42.4\% on the CodeContests \cite{Li_2022} test set. In addition, we found that GPT-turbo-3.5 performs significantly worse at \textit{Analysis-Level} explanations compared to GPT-4. Both of them can generate high-quality \textit{Description}s.

The main contributions of this work are:
\begin{enumerate}
\setlength\itemsep{-1pt}
    \item We advocate disentangling reasoning about the problem and code generation in solving competitive-level programming problems. 
    \item We propose a Specific-to-General prompt to automatically generate structured natural language explanations for <problem, solution> pairs. 
    \item We demonstrate that this proposed method yields convincing explanations that are positively scored by the program authors and serve as effective hints to better solve the problems. 
\end{enumerate}
Though the main focus of this paper is not solving competitive-level programming problems. We further discuss how such explanations can potentially be used to improve problem-solving, which we leave as a potential avenue for future research. 

\section{Background}
\subsection{Challenges in Solving and Annotation}
Competitive-level programming problems \cite{Mirzayanov2020CodeforcesAA} are more indirect compared to many code implementation tasks. Reasoning and problem-solving strategies are usually necessary before implementation  \cite{Skiena2003ProgrammingCT, laaksonen2020guide}; and they require solutions to be both correct and efficient. While brute-force solutions may be feasible for some problems, they are frequently deemed inadequate due to their high time and space complexity. Additionally, some problems may intentionally obscure the key idea behind the solution, presenting more puzzle-like challenges. 

The challenges in solving competitive-level programming lie in not only the implementation phase but also the reasoning process that precedes it, which has not been adequately addressed by previous works. Consider the problem in Table \ref{tab:examples}, if given a specific instruction, LLMs optimized for code generation can generate the correct program. 
However, the reasoning process of why it is correct is not reflected in the problem or the program solution. To bridge the gap between the problem and the solution, natural-language-explained solutions and reasoning steps can be potentially helpful.

Annotating explanations on how to solve those questions can be difficult and time-consuming, even for highly skilled programming competitors. Solutions are written under a time constraint, and competitors compromise readability for fast implementation. Therefore, solutions are often hard to understand by others without natural language explanation. Small-scale solution explanations, also known as editorials, can be found in some blogs on the Internet, but collecting large-scale editorials in a unified format is still infeasible. 
In this paper, we tackle how to use LLMs to generate silver-standard explanations automatically, thereby addressing the need for accessible and comprehensive solution explanations in competitive-level programming.

\subsection{Problem Formulation}
We formalize our task with a problem set consisting of $n$ problems $P=\{p_1, p_2,\cdots,p_n\}$; each problem $p_i$ is a text sequence that describes the following aspects clearly. 
\begin{itemize}
\setlength\itemsep{-1pt}
    \item Problem statement: a natural language description of the problem, as the first cell in Table \ref{tab:examples}.
    \item Input/Output: the format and input/output constraints (e.g. ranges) for the submitted program $s$.
    \item Example: An example of a correct pair of input/output.
    \item Note (Optional): Explanation of the Example input/output.
\end{itemize}
  
Each $p_i$ corresponds to a set of oracle human solutions $S_i=\{s_i^1, s_i^2, \cdots, s_i^t\}$ where $t$ is the number of total solutions of $p_i$, we then select top $k$ solutions following 2 simple rules: (1) We only consider correct human solutions, i.e., those that have passed the online judge system; (2) Solutions in $S_i$ are ranked according to their programming language and size in bytes, with a preference for Python-implemented solutions and shorter programs.

All experiments in this paper are \textbf{zero-shot} on large language models without fine-tuning. 

\subsection{General-to-Specific Prompting Solver}
Before delving into the explanations, we first discuss the general capacity of LLMs to solve those problems directly from problem to generated solution or thinking step-by-step. We note that our methodology requires using instruction-finetuned language models \cite{ouyang2022training}, as we provide zero-shot instructions in our prompt.

\begin{figure}
	\centering
\includegraphics[width=0.45\textwidth]{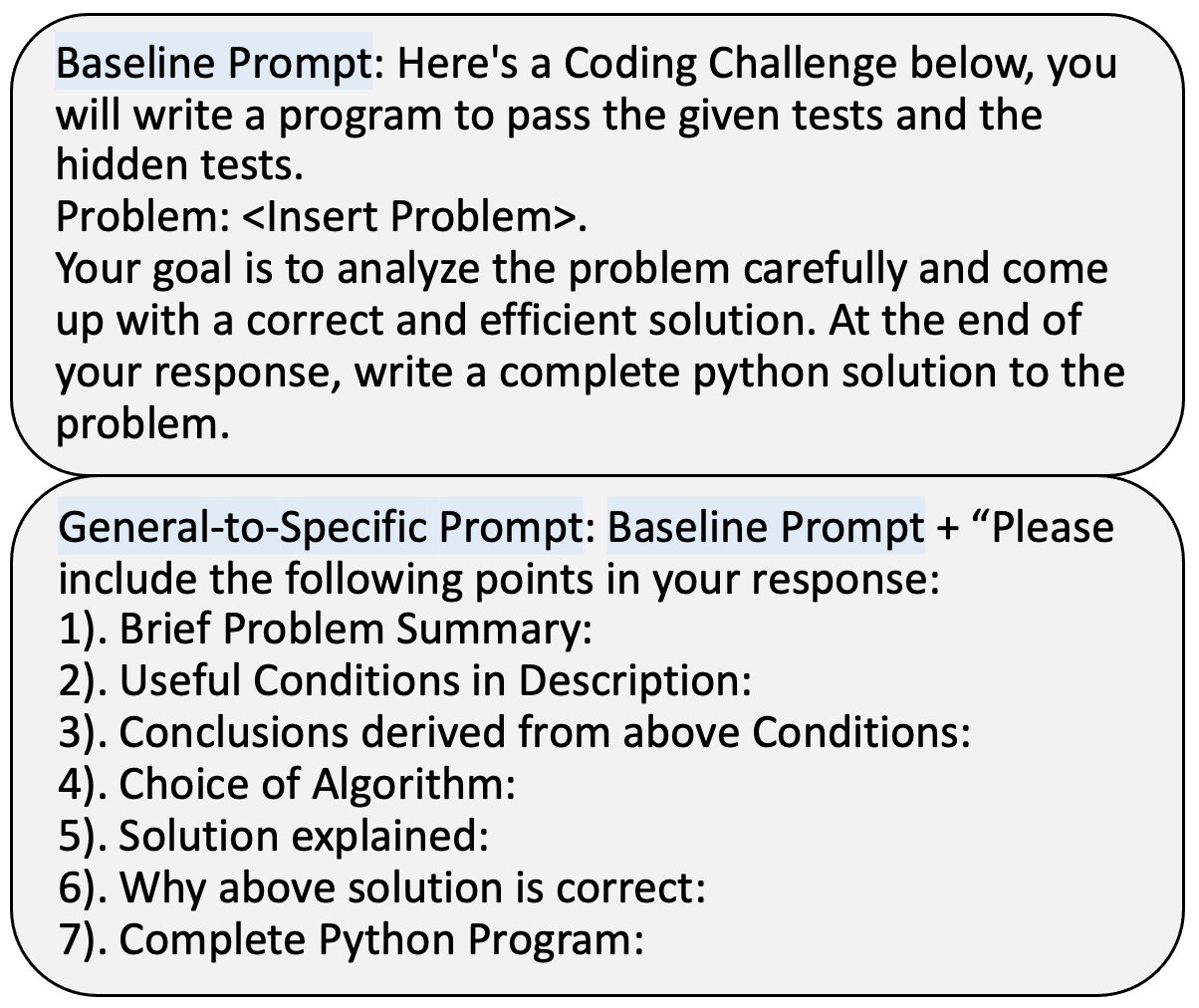}
	\caption{The Baseline Solver Prompt and General-to-Specific (G2S) Prompt which asks LLMs to follow the reasoning steps till it reaches the state of implementation. }
	\label{fig:cot}
\end{figure}

We designed a general-to-specific reasoning chain, which is inspired by humans' step-by-step thought processes in solving such problems. As shown in Figure \ref{fig:cot}, we prompt the LLM to start from a \textbf{general} understanding of the problem and potential algorithms to use, then gradually transit to a more \textbf{specific} and detailed level of understanding, till finally implementing a program in Python.  

For each problem, we generate $k$ programs $\{ {g_i^1},{g_i^2},\cdots, {g_i^k} \}$ with LLMs as the $k$ candidates to conduct a solve@k evaluation, as defined by \citet{chen2021evaluating}. In other words, if any of the generated $k$ programs is considered as a correct solution, then this problem is regarded as solved.

When experimenting with GPT-turbo-3.5 on the 165 problems in the test set of CodeContests, the proposed general-to-specific prompt can boost the solve@10 
from 6.1\% to 9.1\%. Through reasoning general-to-specific, the LLM can perform a bit better at solving programming problems. However, upon examining the failed cases, we discovered that for most problems, the model makes a mistake at a very early stage, ultimately resulting in a completely incorrect solution. 

\section{Method}

In the process of problem-solving, a human typically constructs a solution by progressing from a general idea to a detailed code implementation. However, explaining that solution involves a reverse approach. This entails examining the code on a line-by-line basis, interpreting the role of each function, and then rationalizing the algorithmic steps in relation to the original problem. Therefore, we design a specific-to-general explanation generation method.

\subsection{Specific-to-General Solution Explaining}
Previous works have demonstrated the ability of LLMs to explain code; therefore, we investigated generating explanations automatically using an LLM with both the problem and sample solution as input. 
For a problem-solution pair $\{p_i, s_i^j\}$ where $j\leq k$, an explanation $e_i^j$ is generated. For each problem $p_i$, a set of explanations $E_i$ is generated given different solutions $\{s_i^1, s_i^2, \cdots, s_i^k\}$. 

Although simple prompts such as 'explain the solution' may generate useful explanations, these often lack crucial information and are difficult to evaluate due to their output's diversity. To tackle this issue, we deliberately control aspects of the explanations, requiring them to include a 'problem summary' that demonstrates our understanding of the problem and three levels of 'natural language description of the problem,' illustrating our ability to comprehend the solution from a low-level to a high-level perspective. These can be considered as 'Description-level' explanations. The elements such as 'used algorithm,' 'time complexity,' and 'proof of correctness' fall under 'Analysis-level' explanations, showcasing the language model's overall analysis and understanding of the solution. The method for this specific-to-general explanation prompt is detailed in the left part of \ref{fig:main method}.

Format-guided-generated explanations are clear and structured, thus making it easier to disentangle information and evaluate each aspect. In our experiment, over 99\% of explanations contain all defined points, with less than 1\% skipping some later points due to the length constraint. 

In addition, thinking from detailed code-level implementation can also provide the intermediate steps in context. The LLM can reach a better general understanding of the solution by looking at its previously generated general descriptions. 

\subsection{Explanation Instructed Solver}
In order to evaluate the quality of generated explanations, we design an automatic metric to test how much it can aid in solving the problem if included in the instruction. In this setting, we give both the original problem as well as one of \textit{Description-level} points to the LLM Solver with the corresponding prompt given in the right part of Figure \ref{fig:main method}. If a given instruction enables the LLM Solver to solve a problem it was previously unable to solve, we consider that instruction to be more informative than one that does not yield such an outcome.

 \section{Experiments}
\subsection{Experimental Setup}
\paragraph{Model}We use both GPT-3.5-turbo and GPT-4~\cite{oai2023chatgpt,openai2023gpt4} as the Explainer for explanation generation.\footnote{Due to the usage limit of GPT-4, we run larger scale experiments only on GPT-3.5-turbo.} We use GPT-3.5-turbo for all our experiments as Solver LLM for code generation. We will refer to it as GPT-3.5 for simplicity. The temperature $t$ is set to $0$ wherever only one sample is needed, and $0.2$ otherwise. Max-length of text is set to 4096, and we skipped $0.7\%$ of cases where the max length is exceeded.

\paragraph{Data}To ensure the effectiveness and accuracy of our results, given that GPT-3.5 may have seen some <problem, solution> pairs in its training data, we use the CodeContests test set as our main dataset in this paper. It contains 165 real online contest problems from \textit{Codeforces}, the earliest of which dates back to Oct 2021, which is after the knowledge cutoff of GPT-3.5 and GPT-4 (Sep.\ 2021). Additionally, we extract a small subset of 50 more recent problems from \textit{Codeforces} for human evaluation.  Table \ref{tab:dataset} are statistics based on their level-of-difficulty ratings. Problems with ratings over 2k are considered very difficult, most of which can only be solved by medal-winning competitors.

\begin{table}[h]
\centering
\scalebox{0.67}{
\begin{tabular}{l|c|c|c|c|c}\hline
& \multicolumn{5}{c}{Ratings} \\ \hline
Dataset & total & \small{[800, 1000]} &\small{(1000, 1500]}& \small{(1500, 2000]} & \small{(2000, 3600]}\\ \hline
\small{CodeContests} & $165$ &$18.2\%$ & $17.0\%$ & $20.0\%$ & $44.8\%$\\ 
\small{Our Data} & $50$ & $34\%$ & $46\%$ & $20\%$ & $0\%$\\ \hline
\end{tabular}}
\caption{Difficulty statistics (higher ratings = more difficult) for the dataset. The problems in our dataset exclude hard problems (rating over 2k), as they exceed the rating of our annotators.}
\label{tab:dataset} 
\end{table}

\paragraph{Metric} We employ pass@k \cite{chen2021evaluating} as our evaluation metric for solve rate. For each problem $p_i$, we sample $k$ programs generated from GPT-3.5 and evaluate them using \textbf{Solve Rate@k} metric: the percentage of programs that pass all hidden test cases when submitted to Codeforces' online judge.
We first filter the programs by their output on the public test cases before submitting them and also measure \textbf{Pass Public@k}: the percentage of programs that pass the public test cases given in the examples. The above metrics are abbreviated as `solve@k' and `public@k'.

\subsection{Human Evaluation}
We measured the quality of LLM-generated explanations using human evaluation. We collect 50 \textit{<problem, solution>} pairs from Codeforces, ensuring that their format remained consistent with those in CodeContests.

\paragraph{Author Likert Scores} Recognizing that understanding and explaining others' solutions can be a challenging task for programmers, we employed an annotator-centered evaluation approach. We extracted solutions and corresponding problems from Codeforces for an expert annotator. The Explainer then generates an explanation for the annotator's solution, which was subsequently scored by the author of the explained solution. Note that each explanation is scored by the author of the solution being explained. 

We generated explanations for 50 problems with ratings ranging from 800 to 2000, along with their corresponding solutions, and provided these explanations to human experts. They were asked to assign a Likert score from $-2$ (very poor) to $2$ (excellent). 

\begin{figure}
	\centering
\includegraphics[width=0.49\textwidth]{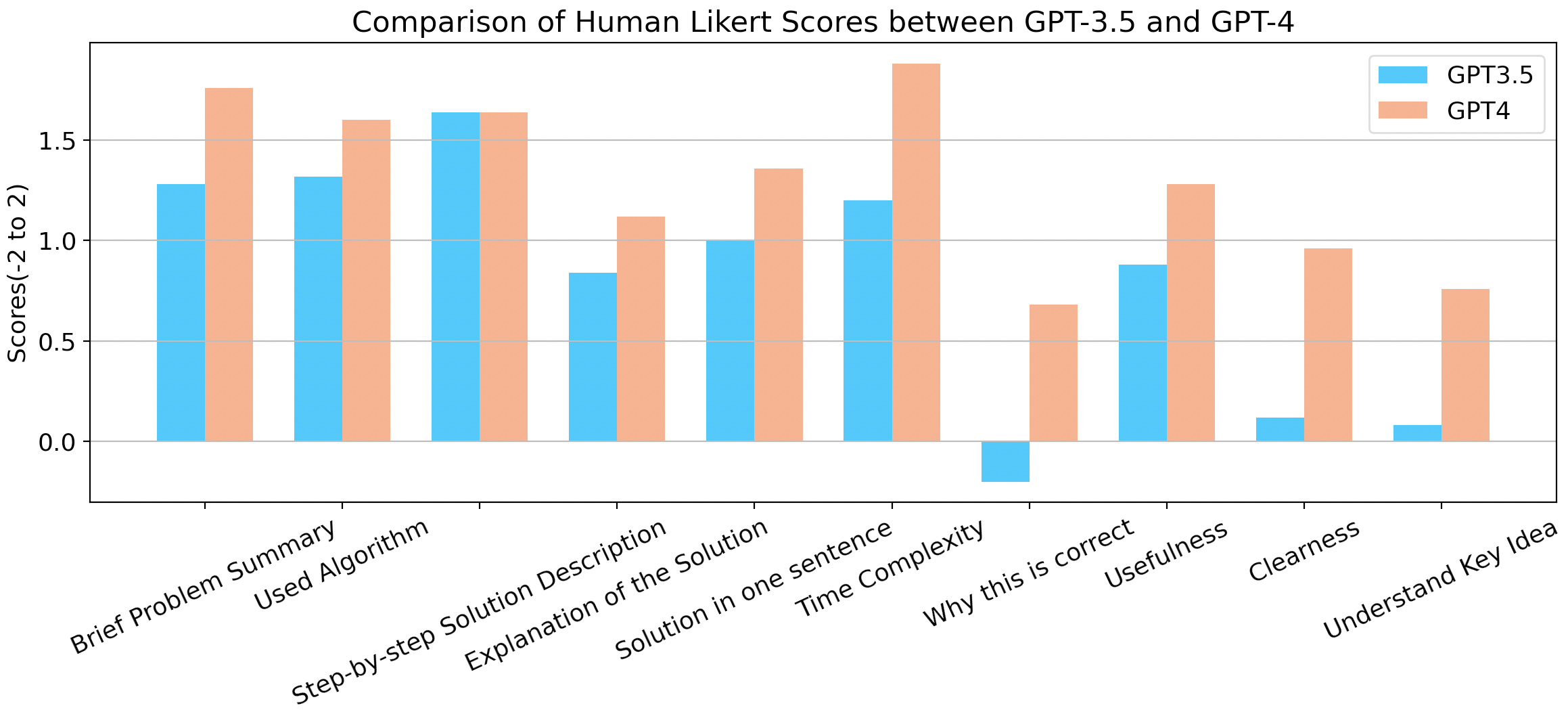}
	\caption{Human Likert scores ($-2$: very poor to $2$: excellent)  evaluating various aspects of the explanations. }
	\label{fig:human}
\end{figure}
The evaluation consists of ten questions, each one corresponding to a specific aspect of the explanation. We separately assess the quality of the response to each point of our G2S prompt (see Figure \ref{fig:main method}). Furthermore, we developed three criteria to evaluate various aspects of the overall explanation:
\begin{enumerate}
\setlength\itemsep{-1pt}
    \item \textit{Usefulness}: How useful is the explanation as guidance to solve the problem? 
    \item \textit{Clearness}: How good is the explanation in terms of describing everything clearly and avoiding ambiguity?
    \item \textit{Understanding}: How much does the LLM understand the key idea behind the solution?
\end{enumerate} 
The average Likert scores over 50 problems are shown in Figure \ref{fig:human}. Regarding the scores for the solution descriptions (\textit{Step-by-Step Solution Description, Explanation of the Solution, Solution in One Sentence}) and usefulness, both GPT-3.5 and GPT-4 Explainer are positively rated by expert annotators, with an average of $1.16$ and $1.36$ respectively. 

However, GPT-3.5 receives near zero or negative scores on questions including \textit{why it's correct, clearness, and understanding}, showing its inadequate ability to grasp the key idea behind the solution, while GPT-4 performs better ($0.68\sim0.88$ score higher) on these aspects. This reveals a clear difference in the abilities of GPT-3.5 and GPT-4 to reason and analyze competitive-level programming solutions. 

\paragraph{Qualitative Analysis} 
We observed several interesting aspects of the explanations generated by the models. Models can effectively explain code by integrating the problem statement and the solution. Step-by-step descriptions are often more concise than line-by-line code reports by summarizing complex operations and avoiding re-stating well-known algorithms (e.g., depth-first-search). 

A sample explanation from GPT-3.5 is given in Table \ref{tab:qualitative}. It describes the solution very well in both specific (step-by-step) and general (one-sentence) levels. It summarizes the operations of `count $<0$' and `multiply -1 or 1' into  `negative on the left, positive on the right' and explains if it's sorted, then `yes' otherwise, `no'. However, if we look at the one-sentence description, there are ambiguous terms like `original array' or `move elements', which might mislead the problem-solving if interpreted incorrectly. This is due to natural languages' ambiguous nature compared to programming languages. 

Models exhibit shortcomings when explaining solution correctness, as they may not comprehensively account for conditions stipulated in the problem statement. For instance, when explaining Example \ref{tab:examples}, it failed to recognize that ``swapping signs of 2 elements multiple times means moving signs arbitrarily along the array'' is a crucial condition, which is not mentioned explicitly in natural language. This highlights a potential limitation in the models' ability to extract and incorporate essential information from various parts of the problem statement when generating explanations.

We also present the full input/output and our scores for both successful and failed cases in appendix \ref{sec:appendix}.

\begin{table}[ht]
\small
  \centering
    \begin{tabular}{|p{7.5cm}|}
    \hline
   \textbf{1). Brief Problem Summary:} 
     Given an array of n integers, check if it can be sorted by swapping the signs of any two elements with different signs. \\ \hline
    \textbf{2). Used Algorithm:}  
    The algorithm used is a simple implementation of the problem statement.\\ \hline
    \textbf{3). Step-by-step Solution Description:} \\
1. Read the input array a. \\
2. Count the number of negative elements in the array and store it in h. \\
3. Create a new array b by {\color{cyan}taking the absolute value of each element in a and multiplying it by -1 if the index is less than h, otherwise by 1.} \\
4. Check if the sorted array b is equal to the original array b. \\
5. If they are equal, print "yes", otherwise print "no".\\
    \textbf{4). Explanation of the Solution:} The solution works by first counting the number of negative elements in the array. We then create a new array b by taking the absolute value of each element in a and multiplying it by -1 if the index is less than h, otherwise by 1. {\color{cyan}This ensures that all negative elements are on the left side of the array and all positive elements are on the right side.} We then check if the sorted array b is equal to the original array b. If they are equal, it means that we can sort the original array by swapping the signs of any two elements with different signs. \\ \hline
\textbf{5). Solution in one sentence:} The solution works by counting the number of negative elements in the array, creating a new array with {\color{cyan}all negative elements on the left and positive elements on the right}, and checking if the sorted array is equal to the original array. \\ \hline
\textbf{6). Time Complexity:} The time complexity of the solution is O(nlogn) due to the sorting operation.\\ \hline
\textbf{7). Proof of correctness (Why this is correct):} The solution is correct because it ensures that all negative elements are on the left side of the array and all positive elements are on the right side. This means that we can sort the array by swapping the signs of any two elements with different signs. {\color{red}The sorted array will be equal to the original array if and only if we can sort the array using the given operation.}\\ \hline
  \end{tabular}
  \caption{GPT-3.5's explanation to the example in Table \ref{tab:examples}, which understands and clearly describes the key idea behind the solution. Note it comprehends the code from a detailed level in `3)' and a general level in `5)'. blue: correct, red: incorrect)}
  \label{tab:qualitative}
\end{table}

\subsection{Automatic Metrics: Improving Solutions}

We further investigated the ability of generated explanations to improve problem solving. Our fundamental assumption is that if an explanation accurately describes the algorithm, it should be capable of guiding the implementation of a correct solution. Consequently, we experimented with versions of the Instructed Solver Prompt in Figure \ref{fig:main method}, wherein one point in the explanation (i.e., an aspect of the solution) is provided to the GPT-3.5 Solver as a \textit{hint} for solving the problem. 

We compare it with two baseline solvers that, unlike our solver from Figure \ref{fig:main method}, are not conditioned on explanations and only get the problem statement as an input: zero-shot prompt (denoted as Baseline in Table \ref{tab:main-eval}) and General to Specific (G2S) ``step-by-step'' prompt shown in Figure \ref{fig:cot}. We also check that explanations do not contain code snippets to ensure the solutions are not directly leaked in explanations. However, note that it is still not a completely ``fair'' comparison, since the automatically generated `silver explanations' are conditioned on oracle solutions. 

\paragraph{Main results}  For  GPT-3.5, we measure pass@k  for $k=\{1,5,10\}$, but only pass@1 for GPT-4 due to access limits. To sample $k$ programs, we sample $k$ different human solutions for Explainer and then generate a program for each explanation. 

Results are shown in Table \ref{tab:main-eval}. Different {\it Description-level} aspects of explanations improve both the solve rate and pass public rate. The most detailed aspect, \textit{Step-by-Step Solution Description} (S-by-S), which provides a detailed natural language description of the implementation, offers the most significant benefit to problem-solving, resulting in a solve rate @1 that is 7.4 times higher than the baseline. The impact of \textit{Explanation of the Solution} (Exp-Sol) and \textit{Solution in One Sentence} (OneSent) is comparatively lower due to their concise nature, which offers a less clear path towards the solution. However, providing information on the algorithms used (UsedAlg) or the expected time complexity (TC) does not improve GPT-3.5's problem-solving capabilities.

The pass@1 results for GPT-4 Explainer are not significantly better than for GPT-3.5, indicating that they share similar capabilities in terms of \textit{Description-level} explanations. 

\begin{table}[h]
\centering
\scalebox{0.76}{
\begin{tabular}{l|c|c|c|c}\hline
\multicolumn{5}{c}{GPT-3.5 Solver}\\ \hline
& {solve@1} &{solve@5}& solve@10 &public@10\\ \hline
Baseline & $1.8$ &$3.6$ & $6.1$ & $13.9$\\ 
G2S prompt & $2.4$ &$ 5.4$ & $9.1$ & $18.8$\\ \hline
\multicolumn{5}{c}{GPT-3.5 Solver With Silver Explanation}\\ \hline
w/ UsedAlg & $1.8~\textit{(1.2)}$ &$4.2$ &$6.1$ & $13.3$\\
w/ S-by-S & $13.3~\textit{(15.8)}$ &$32.2$ & $42.4$ & $47.9$ \\
w/ Exp-Sol &$6.1~\textit{(4.8)}$ & $17.6$ & $23.6$ & $32.7$\\
w/ OneSent & $4.2~\textit{(4.2)}$ & $9.1$ &$13.9$ & $26.1$\\
w/ TC($O(\cdot)$) &$1.8~\textit{(2.4)}$ & $3.6$ & $6.7$ & $13.3$\\\hline
\end{tabular}}
\caption{Different aspects of the explanation's effect on improving program generation. Values are percentage \% and `solve' and `public' are short for `Solve Rate' and `Pass Public Tests'. Solve@1 results in \textit{parentheses} are from GPT-4's generated explanations. The bottom 5 rows correspond to Figure \ref{fig:main method}'s points 2,3,4,5, and 6 in the left prompt.} 
\label{tab:main-eval} 
\end{table}
\paragraph{Pass Public Tests vs. Solve} One observation from Table \ref{tab:main-eval} is that solve@10 is significantly less than public@10. For a program that passes the public tests but fails the hidden tests, there are two possibilities: 1) It is incorrect and only applies to a subset of test data, including the public tests; 2) It is inefficient. As discussed before, in competitive-level programming, a ``correct'' but slow implementation does not count as a solution, as there are constraints on time and space complexity. Therefore, we further study programs that pass the public tests but may fail hidden tests. As shown in Table \ref{tab:tle}, the baseline has 48.9\% of its programs rejected by the online judge due to inefficiency, indicating that GPT-3.5 tends to generate inefficient implementations (e.g., brute force solutions). 
\begin{table}[h]
\centering
\scalebox{0.82}{
\begin{tabular}{l|c|c|c|c}\hline
& {Solve} &{Wrong Answer}& TLE & Other\\ \hline
Baseline & $35.1\%$ &$15.6\%$ & $48.9\%$ & $0\%$\\ 
G2S prompt & $38.3\%$ & $14.1\%$ & $47.6\%$ & $0\%$\\ \hline
w/ UsedAlg & $39.1\%$ &$18.9\%$ &$42.1\%$ & $0\%$\\
w/ S-by-S & $75.6\%$ &$11.4\%$ & $11.4\%$ & $1.6\%$ \\
w/ Exp-Sol &$73.6\%$ & $11.9\%$ & $11.1\%$ & $1.4\%$\\
w/ OneSent & $56.6\%$ & $27.9\%$ &$14.0\%$ & $1.5\%$\\\hline
\end{tabular}}
\caption{Final judgement of generated programs that pass the public tests. TLE means time limit exceeded, and \textit{other} includes memory limit exceeded and runtime error.\protect\footnotemark}
\label{tab:tle} 
\end{table}
\footnotetext{This is for all submissions, i.e., one problem might have up to $k$ submissions, which is different from the problem-wise solve rate. }

When provided hints from the solution description, the portion of TLE programs drops significantly. Although GPT-3.5 may still make mistakes in some details or fail to consider corner cases even with hints from the explanation, it is better at avoiding inefficient solutions.

Another interesting observation is that the wrong answer rate for one-sentence explanation-instructed solving is higher than the baseline. One possible explanation is that it is challenging to incorporate corner case handling in a one-sentence solution description, which makes GPT-3.5 more likely to implement an almost-correct program.

\paragraph{Difficulty of the problem} We further study the  aiding effect of three levels of \textit{Solution Description} on problems of different difficulty ratings.
Codeforces problems are given ratings, the higher the ratings are, the more challenging the problem is. Individuals who consistently solve problems with ratings of 2000 are in the 93rd percentile of all participants.
\begin{figure}
	\centering
\includegraphics[width=0.48\textwidth]{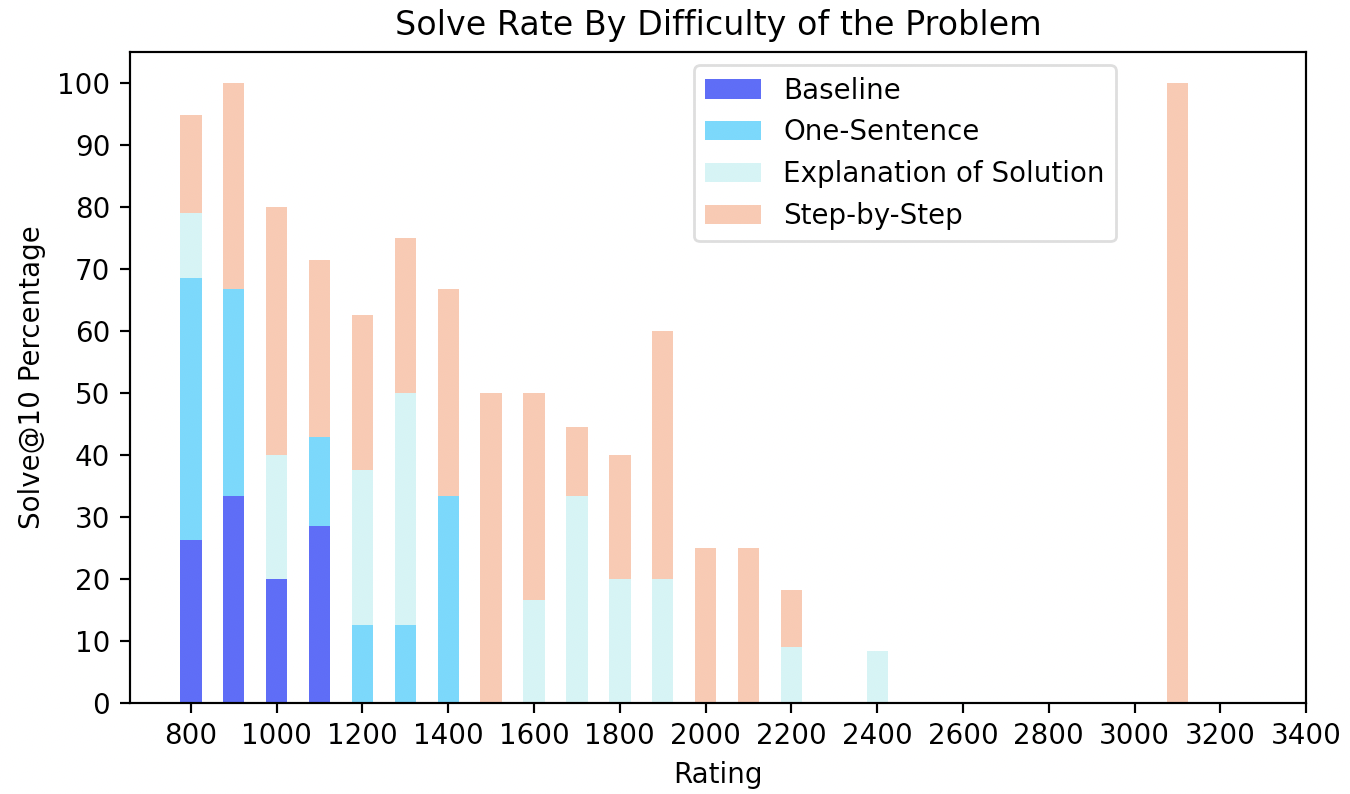}
	\caption{The aiding effects of 3 levels of \textit{Solution Description} over different difficulty ratings. The difference in color shows the gain in solve@10. \protect \footnotemark}
	\label{fig:rating}
\end{figure}
\footnotetext{The outlier is because there's only one rating=3100 problem in the CodeContests test set.}
As shown in Figure \ref{fig:rating}, the solve rate decreases as the ratings increase and no explanation can help solve complex problems. However, for easier problems, even a one-sentence hint enables GPT-3.5 to solve approximately 70\% of problems, compared to the $\sim30\%$ baseline. Furthermore, hints can effectively help to solve medium-difficulty problems which were previously unsolvable. 

\paragraph{Sampling Strategies}
In our approach, we generate $k$ programs and treat all of them as candidates without re-ranking, making the sampling strategy crucial. We therefore compared three strategies for sampling $k$ programs.
\begin{enumerate}
\setlength\itemsep{-1pt}
    \item Sample $k$ human solutions: for each $p_i$, we sample $S_i=\{s_i^1, s_i^2, \cdots, s_i^k\}$, and for each of the solution$s_i^j$, we generate one explanation $e_i^j$, and one corresponding program $g_i^j$.
    \item Sample $k$ explanations: We only take the first solution $s_i$, and sample $E_i=\{e_i^1, e_i^2, \cdots, e_i^k\}$, for each explanation $e_i^j$, we generate one corresponding program $g_i^j$.
    \item Sample $k$ programs: We only sample 1 solution $s_i$ and one corresponding explanation $e_i$, then we sample $k$ programs $G_i=\{g_i^1, g_i^2, \cdots, g_i^k\}$ given the explanations.
\end{enumerate}
\begin{table}[h]
\centering
\scalebox{0.7}{
\begin{tabular}{l|c|c}\hline
w/ OneSent & public@10 &solve@10\\ \hline
~~ Baseline  & $13.9\%$ & $6.1\%$\\ 
Statg1. Sample 10 Human Solution  $s_i$ &$\textbf{26.1\%}$ & $\textbf{13.9\%}$\\
Statg2. Sample 10 Explanation $e_i$  & $24.8\%$ & $12.1\%$ \\
Statg3. Sample 10 Programs $g_i$  & $18.2\%$ & $6.7\%$\\
Statg3. 10 $g_i$ from GPT-4 explanation$^*$ & $18.2\%$ & $10.9\%$\\\hline
\end{tabular}}
\caption{Comparison of sampling strategies. Strategies are numbered. Rows 2,3,4 are GPT-3.5 sampling 10 solutions/explanations/programs respectively, the last row is GPT-4 Explainer sampling 10 programs from a single explanation.}
\label{tab:sample} 
\end{table}

Table \ref{tab:sample} shows that the first strategy of sampling from 10 different human oracle solutions is the most effective. Additionally, the second strategy of sampling 10 explanations from one oracle solution yields better results than sampling 10 programs from one explanation (strategy 3). 
One potential reason is that some human solutions may have poor readability or employ complex implementations that are hard to follow.  By sampling different human oracle solutions, there is a higher likelihood that explanations based on clear and concise solutions can serve as better hints. Similarly, sampling diverse explanations can mitigate the issue of misleading, incorrect explanations. We also compared the explanation quality of GPT-4 (i.e., only as an Explainer) and found it to be superior to GPT-3.5 in the same setting (10.9\% vs. 6.7\%). We skipped other settings due to experimental limitations.

\section{Related work}

\subsection{Solving Competitive-level programming problems}
Early attempts to apply deep learning to solve competitive-level programming problems \cite{balog2017deepcoder} utilized traditional approaches such as SMT solvers and search to generate short programs for simple problems. \citet{DBLP:journals/corr/abs-1802-04335} collected a dataset of human-written problem statements and solutions for Codeforces problems and introduced sequence model baselines that could solve a small subset of their dataset. With the advent of Transformers, AlphaCode \cite{Li_2022} achieved significant progress in solving competitive-level programming problems by attaining a rating equivalent to the top 54\% of participants on Codeforces by finetuning LLMs in the problem-to-solution scenario with the CodeContests dataset collected from Codeforces. Notably, AlphaCode requires sampling 1M program candidates per problem to achieve a 29.6\% solve rate on their test set. \citet{zelikman2023parsel} improves upon AlphaCode by using fewer samples for the same level of performance.   Our study focuses on explaining solutions to problems, rather than directly solving them. To the best of our knowledge, this is the first attempt to explain competitive-level programming solutions using language models, which places a landmark of the reasoning and interpreting ability of those models. 

\subsection{Reasoning with large language models}
\citet{wei2022chain} has demonstrated that by breaking down the reasoning steps through chain-of-thought (CoT) prompting, LLMs are able to solve challenging reasoning problems by following the correct logic step-by-step.
This method, along with majority voting, has led to notable advancements in solving high-school-level mathematical problems \cite{lewkowycz2022solving}.  \citet{kojima2022large} generalize the idea of CoT to zero-shot learning. Another technique that builds upon CoT is the self-consistency decoding strategy \cite{wang2022self}. This approach samples diverse reasoning paths and selects the most consistent answer, which has shown to improve LLMs' performance on complex reasoning tasks by embracing multiple ways of thinking. Additionally, Parsel \cite{zelikman2023parsel} proposed a framework that focuses on enhancing LLMs' hierarchical multi-step reasoning capabilities, particularly for tasks such as generating complex programs.

\subsection{Code Comprehension with LLMs}
Several existing works have explored generating code explanations using LLMs. \citet{macneil2023experiences} integrated LLM-generated code explanations into an interactive e-book on web software development, showing that students found the generated explanations helpful. \citet{leinonen2023comparing} compared LLM-generated explanations with student-created explanations, finding that the LLM-created explanations were easier to understand and more accurate. \citet{chen2023teaching} utilizes self-generated explanations as feedback to its self-debug. In comparison, our work targets explaining competitive-level programming problems, aiming not only to clarify the code implementation but also to point out the key idea behind the solution, its correctness, choice of algorithms, and time complexity. 


\section{Conclusion and Future Work}
In this paper, we propose explaining competitive-level programming solutions using LLMs. Given a problem and  its corresponding human oracle solution given, LLMs can generate structured explanations that are positively scored by human authors. Our evaluation demonstrates that both GPT-3.5 and GPT-4 exhibit reasonable capabilities in generating faithful and clear descriptions, which can guide another LLM to better solve the problem. GPT-4 outperforms GPT-3.5 significantly in analyzing the problem and solution, as well as capturing the key ideas behind the solution.

Our automatic evaluation metric examines an ideal scenario: when a hint is based on an oracle human solution, it effectively guides the LLM to generate improved programs for solving problems. However, a system should be able to learn from human programming solutions to improve its own problem-solving on novel problems without guidance from a human solution. This raises the question: Can the LLM-generated explanations be utilized to improve subsequent problem-solving? 

Our explanation method can potentially be applied to annotate large-scale data (e.g., the full CodeContests training set), yielding thousands of silver explanations that can be used to fine-tune a reasoning model for competitive-level programming problems. This approach could help bridge the long-standing reasoning gap between problem and program for complex programming problems. Moving forward, we aim to further address solving such problems by focusing on enhancing reasoning for programming problems.

\section*{Limitations}
One primary limitation of this work is that we experimented on only one dataset and two LLMs, namely GPT-3.5 and GPT-4, so it's unclear whether our method can generalize well to other LLMs or problem sources other than Codeforces. Here we just assume that the competitive-level programming problems are well defined so the distribution shift won't be large between sources. 

Another limitation stems from the annotator-centered nature of our human evaluation process, which prevents us from assessing annotator agreement. Individual annotators were only able to score explanations based on their own solutions. While we provided guidelines for assigning scores, the evaluation process remains inherently subjective, and interpretations may vary among different annotators.


\section*{Ethics Statement}
Our research is driven by the potential benefits of improved problem-solving capabilities and a deeper understanding of programming concepts for developers and learners. However, we acknowledge the ethical implications and potential risks specific to our work.

This work focuses on the task of automatic code generation, but we emphasize that it is not intended to replace human efforts in programming. Machine-generated programs may contain errors or vulnerabilities, and it is crucial to thoroughly verify any AI-generated code snippets before using them. Providing code explanations should not be seen as an endorsement to blindly trust the generated programs. Users must carefully understand, verify, and examine AI-generated code to ensure its correctness and safety.
\section*{Acknowledgement}
This material is based on research that is supported in part by the Air Force Research Laboratory (AFRL) and DARPA, for the KAIROS program under agreement number FA8750-19-2-1003. The views and conclusions contained herein are those of the authors and should not be interpreted as necessarily representing the official policies, either expressed or implied, of DARPA, IARPA, NSF, or the U.S. Government. 

We sincerely thank our annotators, especially \textit{smax}, for their efforts in this work, and also all reviewers for their valuable suggestions on this paper. 
\bibliography{anthology,custom}
\bibliographystyle{acl_natbib}

\appendix

\newpage
\section{Appendix}
\label{sec:appendix}

\paragraph{Case Study}
Table \ref{tab:case_study} presents an example of the input/output of the model, which contains our specific-to-general prompt as well as the comparison between GPT3.5 and GPT4 generated explanations. We can see that both GPT-3.5 and GPT-4 describe the problem and solution very well. Both models correspond the solution logic to the problem situation correctly. In the analysis of why this solution is correct, GPT-3.5 and GPT-4 mention the key idea of increasing IQ backward means saving IQ for the future. 

Table \ref{tab:error_analysis} presents an incorrectly explained example. LLMs can describe the problem and illustrate the printing array operation as ``moving the last element to the head of the array''. However, both of them fail to understand the purpose of placing the maximum element in the front as they ignore one crucial condition in the problem: all elements are strictly positive. Nevertheless, GPT-4 maintains a better understanding of the problem by noticing the condition that the given array is already sorted in non-decreasing order. 

\begin{table*}[ht]
\tiny
\centering
\begin{tabular}{|p{0.67\textwidth}|p{0.33\textwidth}|}
\hline
\begin{minipage}[t]{\linewidth}
  \begin{tabular}{@{}p{\linewidth}@{}}
\textbf{Input} You are required to read and try to understand a competitive programming problem statement and interpret its correct solution. Think carefully about why doing this can correctly solve the problem.\\ 
-----------\\ 
Problem:Doremy is asked to test $n$ contests. Contest $i$ can only be tested on day $i$. The difficulty of contest $i$ is $a_i$. Initially, Doremy's IQ is q. On day $i$ Doremy will choose whether to test contest $i$ or not. She can only test a contest if her current IQ is strictly greater than 0.\\ 
If Doremy chooses to test contest $i$ on day $i$, the following happens:\\ 
if $a_i>q$, Doremy will feel she is not wise enough, so q decreases by 1;  otherwise, nothing changes.\\ 
Doremy wants to test as many contests as possible. Please give Doremy a solution.\\ 
Input\\ 
The input consists of multiple test cases. The first line contains a single integer $t (1\leq t\leq 10^4) $— the number of test cases. The description of the test cases follows.\\ 
The first line contains two integers $n$ and $q$ $(1 \leq n \leq 10^5, 1 \leq q \leq 10^9)$ — the number of contests and Doremy's IQ in the beginning.\\ 
The second line contains n integers $a_1,a_2,\cdots,a_n (1 \leq a_i \leq 10^9)$ — the difficulty of each contest.\\ 
It is guaranteed that the sum of n over all test cases does not exceed $10^5$.\\ 
Output\\ 
For each test case, you need to output a binary string s, where $s_i=1$ if Doremy should choose to test contest $i$, and $s_i=0$ otherwise. The number of ones in the string should be maximum possible, and she should never test a contest when her IQ is zero or less.\\ 
If there are multiple solutions, you may output any.\\ 
    \begin{tabular}{@{}p{0.5\linewidth}@{}p{0.5\linewidth}@{}}
      \raggedright
Example Input\\ 
5\\ 
1 1\\ 
1\\ 
2 1\\ 
1 2\\ 
3 1\\ 
1 2 1\\ 
4 2\\ 
1 4 3 1\\ 
5 2\\ 
5 1 2 4 3 &
      \raggedright
Example Output\\ 
1\\ 
11\\ 
110\\ 
1110\\ 
01111
    \end{tabular}
Note\\ 
In the first test case, Doremy tests the only contest. Her IQ doesn't decrease.\\ 
In the second test case, Doremy tests both contests. Her IQ decreases by 1 after testing contest 2.\\ 
In the third test case, Doremy tests contest 1 and 2. Her IQ decreases to 0 after testing contest 2, so she can't test contest 3.\\ 
  \end{tabular}
\end{minipage} & 
\begin{minipage}[ht]{\linewidth}
~\\ \textbf{Hint:} Here's an accepted solution from an expert programmer. Can you analyze the code and identify the algorithmic approach used? Please also describe the steps of the algorithm in natural language, explaining what every named variable means, what each step does and how it works. If there is anything non-obvious in the step, explain why it is done. Please notice that in the algorithm, some conclusions might be first deduced from conditions and directly used.\\ 
\begin{lstlisting}[style=python, basicstyle=\tiny]
def solve():
    n, q =map(int,input().split())
    A=list(map(int,input().split()))
    A.reverse()
    q_now = 0
    record = []
    for a in A:
        if a > q_now and q_now <q: # can take off
            record.append(1)
            q_now += 1
        elif a > q_now and q_now >= q:
            record.append(0)
        else: record.append(1)
    print("".join([str(_) for _ in reversed(record)]))

t=int(input())
for i in range(t):
    solve()
\end{lstlisting}
\vspace{3ex}
Analyze the problem and the accepted code of the correct solution. Avoid ambiguity in your description as possible. Each part should be independently interpretable, i.e., each point should be clear to readers who have no access to other points.\\ 
Answer in the following Format: \\ 
1). Brief Problem Summary:\\ 
2). Used Algorithm:\\ 
3). Step-by-step Solution Description:\\ 
4). Explanation of the Solution:\\ 
5). Solution in one sentence:\\ 
6). Time Complexity:\\ 
7). Proof of correctness (Why this is correct):~\\
\end{minipage} \\

\hline
\multicolumn{2}{|c|}{
  \begin{tabular}{p{0.48\textwidth}|p{0.52\textwidth}}
    \begin{minipage}[t]{\linewidth}
    \raggedright 
    ~\\
    \textbf{GPT-3.5's Output: } \\
\textit{1). Brief Problem Summary}:\\ 
Doremy is asked to test $n$ contests. Contest $i$ can only be tested on day $i$. The difficulty of contest $i$ is $a_i$. Initially, Doremy's IQ is q. On day $i$ Doremy will choose whether to test contest $i$ or not. She can only test a contest if her current IQ is strictly greater than 0. Doremy wants to test as many contests as possible.\\  
\textit{2). Used Algorithm}:\\ 
Greedy Algorithm\\ 
\textit{3). Step-by-step Solution Description}:\\ 
Read the number of test cases.\\ 
For each test case:\\ 
  \quad Read the number of contests and Doremy's IQ in the beginning.\\ 
  \quad Read the difficulty of each contest.\\ 
  \quad Reverse the list of difficulties.\\ 
  \quad Initialize a variable $q_{now}$ to 0 and an empty list record.\\ 
  \quad For each difficulty $a$ in the reversed list of difficulties:\\ 
    \quad \quad If $a$ is greater than $q_{now}$ and $q_now$ is less than q, append 1 to the record list and increase $q_{now}$ by 1.\\ 
    \quad \quad If a is greater than $q_{now}$ and $q_{now}$ is greater than or equal to q, append 0 to the record list.\\ 
    \quad\quad If a is less than or equal to $q_{now}$, append 1 to the record list.\\ 
  \quad Print the binary string obtained by reversing the record list.\\ 
\textit{4). Explanation of the Solution}:\\ 
The solution uses a greedy approach to maximize the number of contests Doremy can test. The idea is to {\color{cyan}start from the last contest and work backwards.} For each contest, if its difficulty is greater than Doremy's current IQ and her IQ is less than the maximum allowed IQ, she can choose to {\color{red}take off} and {\color{cyan} decrease her IQ by 1}. {\color{cyan} If the difficulty is greater than her IQ and her IQ is already at the maximum allowed IQ, she cannot take off and must skip the contest}. If the difficulty is less than or equal to her IQ, she can test the contest. By working backwards, we ensure that Doremy tests as many contests as possible while maintaining her IQ above 0.  {\color{purple}\textit{Likert score: 2}}\\ 
\textit{5). Solution in one sentence}:\\ 
The solution uses a greedy approach to maximize the number of contests Doremy can test by working backwards from the last contest and choosing to test or skip each contest based on its difficulty and Doremy's current IQ.\\ 

\textit{6). Time Complexity}:\\ 
The time complexity of the solution is O(n) for each test case, where n is the number of contests.\\ 

\textit{7). Proof of correctness (Why this is correct)}:\\ 
The solution is correct because it uses a greedy approach that always chooses the optimal solution at each step. By working backwards from the last contest, we ensure that Doremy tests as many contests as possible {\color{cyan}while maintaining her IQ above 0}. The solution is optimal because if we skip a contest that we could have tested, {\color{cyan} we lose the opportunity to test it later} and we may not be able to test as many contests as possible. Therefore, the solution is correct and optimal. {\color{purple}\textit{Likert score: 2}}
    \end{minipage} &
    \begin{minipage}[t]{\linewidth}
    \raggedright
    ~\\
    \textbf{GPT-4's Output: } \\
\textit{1). Brief Problem Summary}:\\ 
Doremy wants to test as many contests as possible without her IQ dropping to 0 or less. The input consists of multiple test cases, each containing the number of contests, Doremy's initial IQ, and the difficulty of each contest. The output should be a binary string indicating which contests Doremy should test.\\ 

\textit{2). Used Algorithm}:\\ 
The algorithm is a greedy algorithm that iterates through the contests in reverse order.\\ 

\textit{3). Step-by-step Solution Description}:\\ 
Read the number of test cases $T$.\\ 
For each test case:\\ 
\quad Read the number of contests n and Doremy's initial IQ $q$.\\ 
\quad Read the list of contest difficulties $A$ and reverse it.\\ 
\quad Initialize $q_{now}$ to 0 and create an empty list called record.\\ 
\quad Iterate through the reversed list of contest difficulties $A$:\\ 
   \quad\quad If the current contest difficulty a is greater than $q_{now}$ and $q_{now}$ is less than q, append 1 to the record list and increment $q_{now}$ by 1.\\ 
    \quad\quad If the current contest difficulty a is greater than $q_{now}$ and $q_{now}$ is greater than or equal to q, append 0 to the record list.\\ 
    \quad\quad Otherwise, append 1 to the record list.\\ 
  \quad Print the record list in reverse order as a string.\\ 
\textit{4). Explanation of the Solution}:\\ 
The solution works by {\color{cyan}iterating through the contests in reverse order} and deciding whether Doremy should test the contest or not based on her current IQ ($q_{now}$) and the contest difficulty. If the contest difficulty is greater than her current IQ and her current IQ is less than her initial IQ, she should test the contest and her IQ will increase. If the contest difficulty is greater than her current IQ and her current IQ is greater than or equal to her initial IQ, she should not test the contest. {\color{cyan}Otherwise, she should test the contest as her IQ will not decrease}. ~~~{\color{purple}\textit{Likert score: 2}} \\ 
\textit{5). Solution in one sentence}:\\ 
The solution is a greedy algorithm that iterates through the contests in reverse order and decides whether Doremy should test the contest based on her current IQ and the contest difficulty.\\ 
\textit{6). Time Complexity}:\\ 
The time complexity of the solution is $O(n)$ for each test case, where n is the number of contests.\\ 
\textit{7). Proof of correctness (Why this is correct)}:\\ 
The algorithm is correct because it ensures that Doremy tests as many contests as possible {\color{cyan}without her IQ dropping to 0 or less}. By iterating through the contests in reverse order, the algorithm can make the optimal decision for each contest based on her current IQ and the contest difficulty. If the contest difficulty is greater than her current IQ and her current IQ is less than her initial IQ, she should test the contest and {\color{red}her IQ will increase}, {\color{cyan}allowing her to test more contests in the future}. If the contest difficulty is greater than her current IQ and her current IQ is greater than or equal to her initial IQ, she should not test the contest as her IQ will decrease and she may not be able to test future contests. {\color{cyan}Otherwise, she should test the contest as her IQ will not decrease}, ensuring that she can test as many contests as possible. {\color{purple}\textit{Likert score: 1}} \\ 

    \end{minipage}
    ~\\
  \end{tabular}
} \\

\hline
\end{tabular}

\caption{A full example of Input prompt and <problem, solution> pair and GPT's generated explanation(output). Likert scores are provided for points (4) and (7), all other points are scored 2 for both models as it describes the problem and solution well. Both models get the key idea to work backward and choose to take the contest greedily. Their analysis of when Doremy should take a contest is correct. One minor issue with GPT-4 is that its analysis does not specify that its description(future, increase) is from a backward perspective. (see text in color, blue: correct, red: incorrect). }
\label{tab:case_study}
\end{table*}

\begin{table*}[ht]
\tiny
\centering
\begin{tabular}{|p{0.7\textwidth}|p{0.3\textwidth}|}
\hline
\begin{minipage}[t]{\linewidth}
  \begin{tabular}{@{}p{\linewidth}@{}}
\textbf{Input} \\You are required to read and try to understand a competitive programming problem statement and interpret its correct solution. Think carefully about why doing this can correctly solve the problem.\\ 
-----------\\ 
\textbf{Problem:}\\ 
Problem: An array $a$ is called ugly if it contains at least one element which is equal to the sum of all elements before it. If the array is not ugly, it is beautiful.\\ 
For example: the array [6, 3, 9, 6] is ugly: the element 9 is equal to 6 + 3;  the array [5, 5, 7] is ugly: the element 5 (the second one) is equal to 5;  the array [8, 4, 10, 14] is beautiful: $8 \neq 0, 4\neq8, 10\neq8 + 4, 14\neq8 + 4 + 10$, so there is no element which is equal to the sum of all elements before it.\\ 
You are given an array $a$ such that $1 \leq a_1 \leq a_2 \leq \cdots \leq a_n \leq 100$ You have to reorder the elements of $a$ in such a way that the resulting array is beautiful. Note that you are not allowed to insert new elements or erase existing ones, you can only change the order of elements of $a$. You are allowed to keep the array $a$ unchanged, if it is beautiful.\\ 
Input\\ 
The first line contains one integer $t  (1 \leq t \leq 2000)$ — the number of test cases.\\ 
Each test case consists of two lines. The first line contains one integer $n \; (2 \leq n \leq 50)$. The second line contains n integers $a_1, a_2, \dots, a_n \; (1 \leq a_1 \leq a_2 \leq \cdots \leq a_n \leq 100)$
Output\\ 
For each test case, print the answer as follows:\\ 
if it is impossible to reorder the elements of $a$ in such a way that it becomes beautiful, print NO;  otherwise, in the first line, print YES. In the second line, print n integers — any beautiful array which can be obtained from a by reordering its elements. If there are multiple such arrays, print any of them.\\  
    \begin{tabular}{@{}p{0.5\linewidth}@{}p{0.5\linewidth}@{}}
      \raggedright
Example Input\\ 
4\\ 
4\\ 
3 3 6 6\\ 
2\\ 
10 10\\ 
5\\ 
1 2 3 4 5\\ 
3\\ 
1 4 4  &
      \raggedright
Example Output\\ 
YES\\ 
3 6 3 6\\ 
NO\\ 
YES\\ 
2 4 1 5 3\\ 
YES\\ 
1 4 4\\ 
    \end{tabular}
  \end{tabular}
\end{minipage} & 
\begin{minipage}[ht]{\linewidth}
~\\ \textbf{Hint:} Here's an accepted solution from an expert programmer. Can you analyze the code and identify the algorithmic approach used? Please also describe the steps of the algorithm in natural language, explaining what every named variable means, what each step does and how it works. If there is anything non-obvious in the step, explain why it is done. Please notice that in the algorithm, some conclusions might be first deduced from conditions and directly used.\\ 
\begin{lstlisting}[style=python, basicstyle=\tiny]
t = int(input())
for _ in range(t):
    __ = input()
    a = list(map(int, input().split()))
    if a[0] == a[-1]:
        print("NO")
    else:
        print("YES")
        print(a[-1], end=" ")
        print(*a[:-1])
\end{lstlisting}
\vspace{3ex}
Analyze the problem and the accepted code of the correct solution. Avoid ambiguity in your description as possible. Each part should be independently interpretable, i.e., each point should be clear to readers who have no access to other points.\\ 
Answer in the following Format: \\ 
1). Brief Problem Summary:\\ 
2). Used Algorithm:\\ 
3). Step-by-step Solution Description:\\ 
4). Explanation of the Solution:\\ 
5). Solution in one sentence:\\ 
6). Time Complexity:\\ 
7). Proof of correctness (Why this is correct):~\\
\end{minipage} \\

\hline
\multicolumn{2}{|c|}{
  \begin{tabular}{p{0.46\textwidth}|p{0.54\textwidth}}
    \begin{minipage}[t]{\linewidth}
    \raggedright 
    ~\\
    \textbf{GPT-3.5's Output: } \\
\textit{1). Brief Problem Summary:} Given an array, reorder its elements in such a way that the resulting array is beautiful (i.e., no element is equal to the sum of all elements before it). If it is impossible to reorder the elements to obtain a beautiful array, print "NO". \\
\textit{2). Used Algorithm}: {\color{red}Greedy algorithm} ~~{\color{purple}\textit{Likert score: -2}}\\
3). Step-by-step Solution Description: \\
a. Read the number of test cases, $t$.\\
b. For each test case:\\
\quad i. Read the size of the array $n$ and the array $a$.\\
\quad ii. Check if the first and last elements of the array are equal. If they are, print "NO" and move to the next test case.\\
\quad iii. Otherwise, print "YES".\\
\quad iv. Reorder the array by moving the last element to the beginning of the array.\\
\quad v. Print the reordered array.\\
\textit{4). Explanation of the Solution}: The solution uses a greedy approach to reorder the array. If the first and last elements of the array are equal, it is impossible to reorder the array to obtain a beautiful array. Otherwise, we can move the last element to the beginning of the array to obtain a beautiful array. {\color{red}\textbf{This is because the last element is not equal to the sum of all elements before it}}, and moving it to the beginning ensures that no element is equal to the sum of all elements before it.~~{\color{purple}\textit{Likert score: -1}}\\
~\\
\textit{5). Solution in One Sentence}: Reorder the array by moving the last element to the beginning of the array to obtain a beautiful array, unless the first and last elements of the array are equal. \\
~\\
\textit{6). Time Complexity}: $O(n)$, where $n$ is the size of the array. \\
~\\
\textit{7). Proof of Correctness}: If the first and last elements of the array are equal, it is impossible to reorder the array to obtain a beautiful array. Otherwise, moving the last element to the beginning of the array ensures that no element is equal to the sum of all elements before it, {\color{red}\textbf{since the last element is not equal to the sum of all elements before it}}. Therefore, the solution is correct. ~~~{\color{purple}\textit{Likert score: -2}}
    \end{minipage} &
    \begin{minipage}[t]{\linewidth}
    \raggedright
    ~\\
    \textbf{GPT-4's Output: } \\
\textit{1). Brief Problem Summary:} The problem requires us to reorder the elements of an array to make it beautiful, which means no element in the array should be equal to the sum of all elements before it. We are given multiple test cases, and for each test case, we must determine if it's possible to reorder the array to make it beautiful or not. \\
\textit{2). Used Algorithm}: {\color{red}Greedy algorithm} ~~~{\color{purple}\textit{Likert score: -2}} \\
3). Step-by-step Solution Description: \\ a. Read the number of test cases, $t$.

b. For each test case:

\quad i. Read the number of elements in the array, $n$.

\quad ii. Read the elements of the array, $a$.

\quad iii. Check if the first and last elements of the array are equal. If they are, print "NO".

\quad iv. If not, print "YES" and print the reordered array by placing the last element of the array first followed by the remaining elements in their original order.\\
\textit{4). Explanation of the Solution}: \textbf{The given solution relies on the fact that if the first and last elements of the sorted array} are equal, it's impossible to reorder the elements in such a way to form a beautiful array. {\color{red}\textbf{This is because the first element will always be equal to the sum of all elements before it, which is an empty sum, i.e., zero.}} So, in this case, we print "NO". If the first and last elements are not equal, we can reorder the array to make it beautiful by placing the last element (the largest element) first and keeping the remaining elements in their original order. {\color{red}\textbf{This ensures that the largest element will never be equal to the sum of all elements before it, as there are no elements before it. Since the array is sorted, the remaining elements will also not satisfy the ugly condition.}} ~~{\color{purple}\textit{Likert score: -2}}\\
\textit{5). Solution in One Sentence}: For each test case, print "NO" if the first and last elements of the \textbf{sorted array} are equal, otherwise print "YES" and reorder the array by placing the largest element first and keeping the remaining elements in their original order. \\
\textit{6). Time Complexity}: $O(t*n)$ where $t$ is the number of test cases and $n$ is the number of elements in the array.\\
\textit{7). Proof of Correctness}: As explained in point 4, the solution is correct because it checks the condition that the first and last elements of the array are equal or not. If they are equal or not. If they are equal, it's impossible to make the array beautiful. If they are not equal, the largest element is placed first, ensuring the ugly condition is not satisfied by any element in the reordered array. {\color{cyan}\textbf{Since the array is sorted}}, 
{\color{red}\textbf{the remaining elements will also not satisfy the ugly condition.}} ~~{\color{purple}\textit{Likert score: -1}}
    \end{minipage}
    ~\\
  \end{tabular}
} \\

\hline
\end{tabular}
%
\caption{The lemon-picking example. Likert scores are provided for points (2),(4) and (7), all other points are scored 2 for both models. Unlike GPT3.5, GPT4 notices that the input array is already sorted in non-decreasing order in the problem statement but it still fails to explain why moving the largest element to the first positions makes the array beautiful by ignoring the crucial condition that all elements are strictly positive. (see text in bold - blue: correct, red: incorrect). }
\label{tab:error_analysis}
\end{table*}

\end{document}